\documentclass{article}

\usepackage[final]{nips_2017}

\usepackage[utf8]{inputenc} 
\usepackage[T1]{fontenc}    
\usepackage{hyperref}       
\usepackage{url}            
\usepackage{booktabs}       
\usepackage{amsfonts}       
\usepackage{nicefrac}       
\usepackage{microtype}      

\usepackage{graphicx}
\usepackage{subcaption}

\title{Accounting for hidden common causes when inferring cause and effect from observational data}

\author{
  David Heckerman\\
  Santa Monica, CA 90402 \\
  \texttt{heckerma@hotmail.com} \\
}

\begin{document}

\maketitle

\begin{abstract}
Identifying causal relationships from observation data is difficult, in large part, due to the presence of hidden common causes.  In some cases, where just the right patterns of conditional independence and dependence lie in the data---for example, Y-structures---it is possible to identify cause and effect.  In other cases, the analyst deliberately makes an uncertain assumption that hidden common causes are absent, and infers putative causal relationships to be tested in a randomized trial.  Here, we consider a third approach, where there are sufficient clues in the data such that hidden common causes can be inferred.  
\end{abstract}

\section*{Example and basic results}

We illustrate the approach with an example fromm genomics.  We consider the task of a genome-wide association study (GWAS), wherein one tries to identify which genetic markers known as single nucleotide polymorphism (SNPs) causally influence some trait of interest ({\em e.g.}, height).  Figure~1a shows a generative model for the task.  
In many cases, the relationship between the causal SNPs and trait is well represented by multiple linear regression (unlike the special cases of dominance and recessiveness that we learn about in high-school biology).  
The hidden common causes of the SNPs (here represented by a single hidden node) often corresponds to family relatedness (close or distant) among the individuals in the cohort.  
A million or more SNPs can be measured, but only a relatively small fraction of them causally influence the trait.  The goal of causal inference is to identify the SNPs that do.  

If there were no hidden common causes of the SNPs, one could distinguish causal from non-causal SNPs by applying univariate linear regression to assess the correlation between a SNP and trait, producing a $P$ value based on, for example, a likelihood ratio test.  The separation of causal and non-causal SNPs won’t be perfect, as some non-causal SNPs will have small $P$ values by chance.  Nonetheless, the distribution of $P$ values among the non-causal SNPs should be uniform (we say the $P$ values are {\em calibrated}), whereas the distribution of $P$ values among the causal SNPs will be highly skewed to small values, allowing for a separation of causal from non-causal SNPs that is often useful in practice.

When family relatedness is present, univariate linear regression fails because non-causal SNPs are correlated with the trait, As seen in Figure~1a, there are d-connecting paths between each non-causal SNP and the trait through the hidden variable.  These so-called {\em spurious associations} clutter the results, leading researchers on expensive and time consuming wild goose chases.  To address this problem, one could perform multiple linear regression conditioning on all causal SNPs.  Unfortunately, we don’t know which SNPs are causal.  Consequently, an approach now commonly used in the genomics community is to condition on all SNPs except for the one being tested for association.  As there can be millions of SNPs in an analysis, L2 regularization is used to attenuate variance.

Experiments with synthetic data (to be described in more detail) show that this approach of conditioning on all SNPs yields calibrated $P$ values across many GWASs with a wide range of realistic values for degree of family relatedness, number of causal SNPs, and the strength of causal influences (Figure~1b).

\begin{figure}
  \centering
  \begin{subfigure}[b]{0.45\linewidth}
    \includegraphics[width=\linewidth]{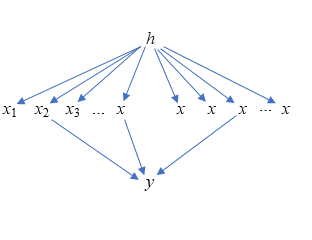}
    \caption{}
  \end{subfigure}
  \ \ \ \ \ \ \ \ 
  \begin{subfigure}[b]{0.45\linewidth}
    \includegraphics[width=\linewidth]{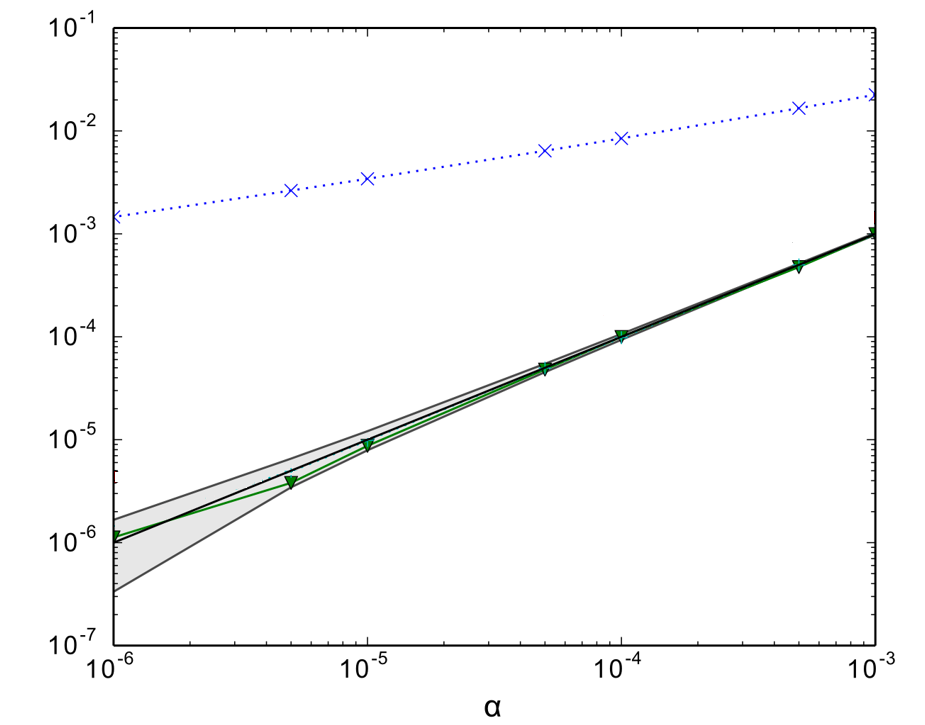}
    \caption{}
  \end{subfigure}
\caption{(a) Graphical model for the data-generation processes in GWAS. The variable $h$ is hidden, representing family relatedness in the cohort.  The variables $x_i$ correspond to SNPs, some of which causally influence the trait $y$.  (b) False-positive rate among non-causal SNPs as a function of the $P$ value threshold $\alpha$ in experiments on synthetic data.  The blue and green lines correspond to univariate regression and L2 regularized multiple linear regression, respectively.  Gray shading represents 95\% confidence intervals assuming $P$ values are calibrated.}
\end{figure}

Now consider a more difficult case shown in Figure~2a, where there is a direct influence of the hidden variable on the trait.  In practice, this influence can happen when different populations or families have different environments that can affect the trait.  In this case, although conditioning on all SNPs does not block all d-connecting paths from non-causal SNPs to the trait, regularized multiple linear regression still yields calibrated $P$ values across a wide range of GWASs (Figure~2b).  

\begin{figure}
  \centering
  \begin{subfigure}[b]{0.45\linewidth}
    \includegraphics[width=\linewidth]{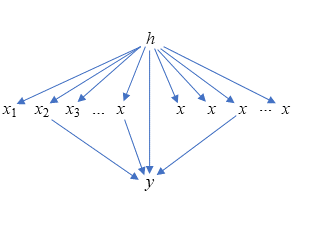}
    \caption{}
  \end{subfigure}
  \ \ \ \ \ \ \ \ 
  \begin{subfigure}[b]{0.45\linewidth}
    \includegraphics[width=\linewidth]{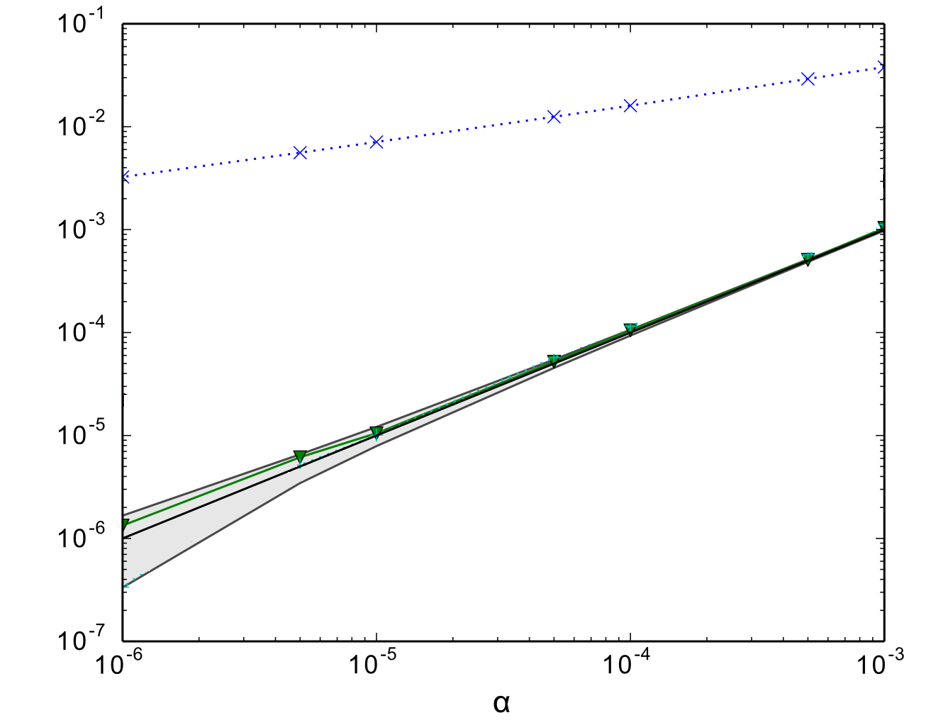}
    \caption{}
  \end{subfigure}
\caption{Figures corresponding to those in Figure~1, but where there is a direct influence of the hidden variable on the trait.}
\end{figure}

Informally, what is happening is that the observation of the many SNPs, all of which depend in a noisy fashion on the same family relatedness, makes it possible to {\em infer} the hidden variable, and thus block the remaining d-connecting paths.  In the remainder of this paper, we examine the data models and these results in more detail.

\section*{L2 regularized multiple linear regression}

Let $y_i$, $x^*_i$, and $\mathbf{X}_{i}=(x_{i1},\ldots,x_{iK})$ denote the trait, test SNP (the one we are computing a $P$ value for), and $K$ remaining SNPs for the $i$th individual, respectively.  For reasons that will become clear, we call the remaining SNPs {\em similarity SNPs.}  Let $\mathbf{y}=(y_1,\ldots,y_N)^{\rm T}$, $\mathbf{x}^*=(x^*_1,\ldots,x^*_N)^{\rm T}$, and $\mathbf{X}=(\mathbf{X}_1^{\rm T},\ldots,\mathbf{X}_N^{\rm T})^{\rm T}$ denote the observations of the trait, test SNP, and $K$ similarity SNPs, respectively, across the individuals.  Thus, $\mathbf{X}$ is an $N \times K$ matrix, where the $ij$th element corresponds to the $j$th similarity SNP of the $i$th individual.  We model the influence of the SNPs on the trait as follows:

\begin{displaymath}
\mathbf{y} \sim \mathcal{N}(\mathbf{1}\mu + \mathbf{x}^*\beta^* + \mathbf{X}\beta; \sigma_e^2\mathbf{I}),
\end{displaymath}

\noindent where $\mu$ is an offset and $\mathbf{1}$ is column of ones, $\beta^*$ is the weight relating the test SNP to the trait, $\beta^{\rm T}=(\beta_1,\ldots,\beta_K)$ are the weights relating the similarity SNPs to the trait, $\sigma_e^2$ is a scalar, and $\mathcal{N}(.; .)$ denotes the multivariate normal distribution.

Using L2 regularization (a Bayesian) approach, we assume that each of the $\beta_i$s corresponding to the similarity SNPs are mutually independent, each having a normal distribution with the same variance

\begin{displaymath}
\beta_i \sim \mathcal{N}(\mathbf{0}; \sigma_g^2), i=1,…,K.
\end{displaymath}

\noindent Further, we standardize the observations of each SNP across the individuals to have variance $1$ (and mean $0$) so that, a priori, each SNP has an equal influence on the trait.

Averaging over the distributions of the $\beta_i$s, we obtain

\begin{equation}
\mathbf{y} \sim \mathcal{N}(\mathbf{1}\mu + \mathbf{x}^*\beta^*; \sigma_e^2 \mathbf{I}+ \sigma_g^2 \mathbf{XX}^{\rm T}).
\end{equation}

\noindent The distribution in (1) is a linear mixed model [1,2].  The distribution also corresponds to a Gaussian process with a linear covariance or kernel function [3].  The model implies that the correlation between the traits of two individuals is related to the dot product of the similarity SNPs for those two individuals, hence the name similarity SNPs.  The similarity matrix $\mathbf{XX}^{\rm T}$ is known as the {\em Realized Relationship Matrix} (RRM) [4].  In general, other similarity measures can be and have been used.  Note that the similarity matrix captures the dependencies among the SNPs induced by the hidden common cause (family relatedness).
 
To compute a $P$ value for the test SNPs, the parameters of the model $(\mu,\beta^*,\sigma_e,\sigma_g)$ are first fit with restricted maximum likelihood.  All parameters can be computed in closed form except the ratio of $\sigma_g^2$ to $\sigma_e^2$, which is usually (and herein) determined via grid search [2].  Then, an F-test is used to evaluate the hypothesis $\beta^*=0$ [5].  To improve computational efficiency with little effect on accuracy, rather than fit $\sigma_g^2/\sigma_e^2$ for each test SNP, we obtain a fit assuming all SNPs are similarity SNPs, and then use it when fitting the remaining parameters for each test SNP [1].

\section*{Experiments}

Both of the experiments described in the opening section are taken from [6].  For each experiment, we generated a large number of GWAS data sets with varying parameters to be described, each with $50,000$ SNPs and $N=4,000$ individuals.  For each data set, we created family relatedness by mating randomly selected synthetic individuals, producing 10 offspring per parent pair. The fraction of offspring in the population was varied across the generated data sets.  In a single mating, the genotype of the child was constructed by selecting one copy of the genotype from the mother and one copy from the father. The SNPs of parents were generated with a minor allele frequency (MAF) sampled uniformly from the range [0.05, 0.5]. Causal SNPs of varying number were then selected at random.  Finally, for each individual, a continuous phenotype was generated from L2 regularized multiple regression on the causal SNPs.  Parameter values used in these simulations were as follows:

\begin{itemize}
\item Fraction of individuals belonging to a family: 0.5, 0.6, 0.7, 0.8, 0.9
\item Number of causal SNPs: 10, 50, 100, 500, 1000
\item $\sigma_g^2/(\sigma_g^2 + \sigma_e^2)$: 0.1, 0.2, 0.3, 0.4, 0.5, 0.6
\end{itemize}

\noindent For each of our two experiments, three data sets for each possible combination of these parameters were generated, yielding 3 x 5 x 5 x 6 = 450 data sets. Different random seeds were used to generate each set of SNPs so that no two sets were the same.  SNPs were generated such that there was no linkage disequilibrium (correlations among SNPs near one another due to meiosis) to simplify the analysis and discussion.

In our first experiment corresponding to no direct arc from $h$ to $y$, we generated $y$ given the SNPs using distribution (1) with no test SNP, causal SNPs $\mathbf{X}$, and $\mu=0$.  In our second set of experiments corresponding to a direct arc from $h$ to $y$, we created that arc by additionally generating 100 hidden causal SNPs drawn from the same family relatedness as the observed SNPs.  That is, we used the generating distribution

\begin{equation}
\mathbf{y} \sim \mathcal{N}(\mathbf{0}; \sigma_e^2 \mathbf{I}+ \sigma_g^2 \mathbf{XX}^{\rm T}
  + \sigma_h^2 \mathbf{WW}^{\rm T}).
\end{equation}

\noindent where $\mathbf{X}$ and $\mathbf{W}$ correspond to the observed and hidden causal SNPs, respectively, and $\sigma_h^2$ is another scale parameter set so that $\sigma_h^2/(\sigma_h^2 + \sigma_e^2) = 0.3$.

For each data set in both experiments, $P$ values were determined using distribution (1) as described in the section on regularized multiple linear regression.

\section*{Discussion}

We can now understand the experimental results.  In the first experiment, although the data was generated using the similarity matrix of the causal SNPs whereas the data was fit using the similarity matrix of all SNPs, $P$ values were calibrated.  Calibration occurred because the two similarity matrices were nearly identical, as they were drawn from the same pattern of family relatedness.

In the second experiment, the similarity matrices of all SNPs, the causal observed SNPs, and the causal hidden SNPs were drawn from the same pattern of family relatedness, and again were nearly identical.  Thus, the fit to the data remained good, and $P$ values were calibrated.  In terms of the causal model, it was possible to infer the family relatedness, in effect inferring $h$ and blocking the $d$-connecting paths in the model.

A closing general remark: GWAS is a very simple problem in causal inference.  We know that SNPs cause traits and not the other way around, so the only real challenge is to identify which SNPs are non-spuriously correlated with the trait.  The fact that this seemingly simple problem requires advanced treatment highlights the complexity of the general problem of causal inference in the presence of hidden causes.

\section*{References}

\small

[1] Yu, J. {\it et al.} A unified mixed-model method for association mapping that accounts for multiple levels of relatedness. {\it Nat. Genet.} {\bf 38}, 203--8 (2006).

[2] Lippert, C. {\it et al.} FaST linear mixed models for genome-wide association studies. {\it Nat. Methods} {\bf 8}, 833--5 (2011).

[3] Rasmussen, C. E. \& Williams, C. K. I.  {\em Gaussian Processes for Machine Learning.}  MIT Press, 2006.

[4] Goddard, M. E., Wray, N., Verbyla, K.\ \& Visscher, P. (2009) {\it Statis. Sci} {\bf 24}:517–529.

[5] Kang, H. M. {\it et al.} Efficient control of population structure in model organism association mapping. {\it Genetics} {\bf 178}, 1709--23 (2008).

[6] Widmer, C. {\it et al.} Further Improvements to Linear Mixed Models for Genome-Wide Association Studies. {\it Sci. Rep.} {\bf 4}, 6874 (2014).

\end{document}